# AttributeNet: Attribute Enhanced Vehicle Re-Identification

Rodolfo Quispe[a,c], Cuiling Lan[b], Wenjun Zeng[b], Helio Pedrini[c]

[a]*Microsoft Corp., One Microsoft Way, Redmond, USA, 98052-6399*
[b]*Microsoft Research Asia, Beijing, China, 100080*
[c]*Institute of Computing, University of Campinas, Brazil, 13083-852*

## Abstract

Vehicle Re-Identification (V-ReID) is a critical task that associates the same vehicle across images from different camera viewpoints. Many works explore attribute clues to enhance V-ReID; however, there is usually a lack of effective interaction between the attribute-related modules and final V-ReID objective. In this work, we propose a new method to efficiently explore discriminative information from vehicle attributes (for instance, color and type). We introduce AttributeNet (ANet) that jointly extracts identity-relevant features and attribute features. We enable the interaction by distilling the ReID-helpful attribute feature and adding it into the general ReID feature to increase the discrimination power. Moreover, we propose a constraint, named Amelioration Constraint (AC), which encourages the feature after adding attribute features onto the general ReID feature to be more discriminative than the original general ReID feature. We validate the effectiveness of our framework on three challenging datasets. Experimental results show that our method achieves the state-of-the-art performance.

## 1. Introduction

Vehicle Re-Identification (V-ReID) aims to match/associate the same vehicle across images. It has many applications for vehicle tracking and retrieval. Like the Person Re-Identification (P-ReID) problem [1, 2], V-ReID has gained increasing attention in the computer vision community [3, 4]. This task is challenging due to drastic changes in view points and illumination, resulting in a small inter-class and large intra-class difference.

Recently, there is a trend to explore additional clues for better V-ReID, such as using semantic maps [5], attributes (such as type, color) [6, 7, 8, 9, 10], viewpoints [11], and vehicle parts [11, 12, 13]. In this work, we focus on the exploration of attributes to enhance the discrimination power of feature representations. Attributes are in general invariant to viewpoint changes and robust to environment alterations.

Most of the previous attribute-based works [10, 9, 7, 8, 6, 14, 15] share a common characteristic in their design: a global feature representation is extracted from an input image using a backbone network (for instance, ResNet [16]), where this feature is followed by two types of heads, one for re-identification (ReID), and the other for attribute recognition. We refer to this design as the Vanilla-Attribute Design (VAD) and illustrate a representative VAD based Network (VAN) in Figure 1. One direct way to use the VAD for V-ReID is to concatenate the embedding features generated from the backbone (that is, global feature) and the attribute-based modules [9, 14, 17].

VAD aims to drive the network to learn features that are discriminative for both V-ReID and attribute recognition, where the attributes are in general invariant to viewpoint and illumination changes. However, there is a lack of effective interaction between the attribute-based branches and V-ReID branch, where the attribute modules learn features for attribute recognition but are not explicitly designed to serve for V-ReID. Wang et al. [8] explore attributes to generate attention masks, but these masks are used only to filter the information from the global feature instead of introducing the rich attribute representation into the final feature representation.

We propose Attribute Net (ANet) to enrich the in-teraction between the attribute features and the V-ReID feature. ANet is designed to distill attribute information and add it into the global representation (from the backbone) to generate more discriminative features. Figures 1 and 2 (with input feature maps obtained from the VAN as illustrated in Figure 1) present the proposed ANet. Particularly, we combine the feature maps of different attribute branches to have a unique and generic representation $G$ of all the attributes.

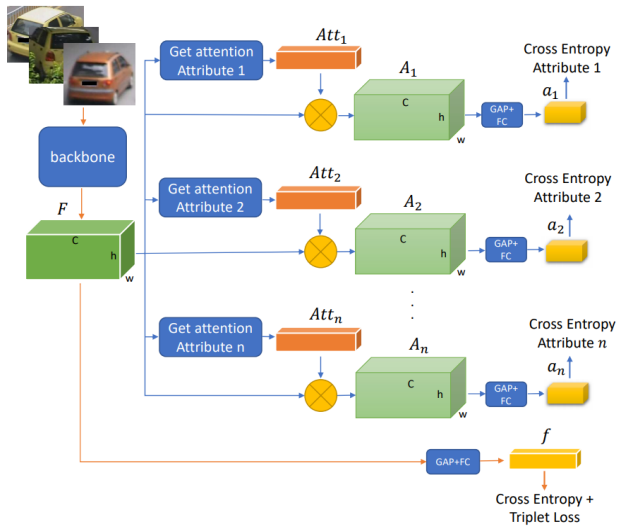

Figure 1: Illustration of VAD based Network (VAN) for V-ReID. It is composed of a backbone network that learns to extract information from an input image and $n$ branches to predict attributes based on attention modules. We use this VAN in our ANet as the first part of our framework.

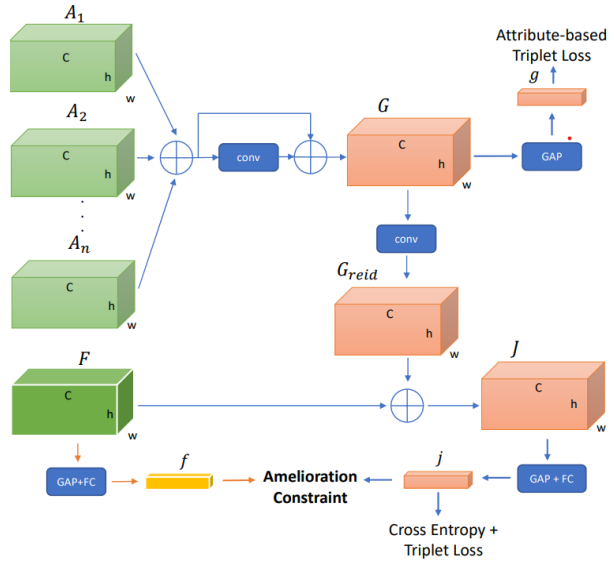

Figure 2: Illustration of the Joint Module. Note that the network to extract feature maps $F, A_1, \cdots, A_n$ is shown in Figure 1 and is not shown here. We distill the helpful attribute feature from $G$ and compensate it onto the global V-ReID feature $F$ to have the final feature map $J$, where the spatial average pooled feature of $J$ is the final ReID feature for matching. Moreover, we introduce a new supervision objective, named Amelioration Constraint (AC) to push $J$ to be more discriminative than the V-ReID feature $F$ before the compensation from attribute feature.

We distill the helpful attribute feature from $G$ and compensate it onto the global V-ReID feature $F$ to have the final feature map $J$, where the spatial average pooled feature of $J$ is the final ReID feature for matching. Moreover, we introduce a new supervision objective, named Amelioration Constraint (AC), to push the compensated V-ReID feature $J$ to be more discriminative than the V-ReID feature $F$ before the compensation from attribute feature.

The main contributions of this work are:

- We propose a new architecture, named ANet, for effective V-ReID, which enhances the interaction between the attribute-supervised modules and V-ReID branch. This drives the distilled attribute features to serve for V-ReID.

- We introduce an Amelioration Constraint (AC), which encourages the attribute compensated feature to be more discriminative than the V-ReID feature before compensation.

Experiments on three challenging datasets demonstrate the effectiveness of our ANet, which outperforms baselines significantly and achieves the state-of-the-art performance.

## 2. Related Work

For vehicle ReID, many approaches explore generative adversarial networks (GANs) [18], graph networks (GNs) [13, 19], semantic parsing (SP) [5] and vehicle part detection (VPD) [20, 12] to improve the performance. Some of them tend to describe the vehicle details [18] and local regions[20, 12]. PRND [20] and PGAN [12] detect predefined regions (such as back mirrors, light and wheels) and describe them with deep features. SAVER [18] modify the input image with the vehicle details erased using a GAN. Then, this synthetic image is combined with the input image to create a new version with the details visually enhanced for ReID. Some works aim to handle the drastic viewpoint changes [13, 5]. Liu et al. [13] describe each vehicle view based on semantic parsing and also encode the spacial relationship between them using GNs.

Vehicle ReID and person ReID share similar purpose in terms of image matching but with respect to the match of different objects (i.e., vehicle and person, respectively). Some approaches have explored attribute information for person ReID. Lin et al. [17] labeled attributes for large datasets and proposed the first deep learning approach using attributes for ReID. Their architecture follows VAD directly by considering a confidence score to re-weigh and create a unique attribute representation and concatenate it with the backbone features. We show that this design is not the most effective way to leverage attributes. Later, other works for person ReID using attributes were proposed to address the imbalance of attribute classes [21], feature and frame disentangle based attributes in video ReID [22], one-shot learning [23] and unsupervised scenarios [24]. However, all of them follow the initial VAD developed by Lin et al. and assume that all the attribute information is equally important for ReID.

For vehicle ReID, similarly, some approaches have explored attribute information [9, 8, 25, 15, 14] or combined attributes with other clues [10, 7, 6]. Most of the previous attribute-based works used attribute information to regularize the feature learning [10, 9, 7, 8, 6, 14, 15]. PAMTRI [7] focuses on solving viewpoint changes using a pose estimation module that learns from keypoints, which is combined with attribute clues to create a final vehicle embedding. StRDAN [10] and DF-CVTC [6] expand the idea of introducing viewpoint-invariant features and used synthetic images during training. They regress the attribute classes from the backbone features, along with the ReID supervision based on the backbone features following VAD. However, using separate heads for different tasks ignores the interaction between the two tasks, where the attribute branches should serve for better ReID. There is a lack of efficient interaction between attribute and ReID information.

Our work addresses this issue by distilling useful attribute information and compensating into the ReID feature representation to have a more discriminative representation.

## 3. Proposed ANet

Our proposed ANet is designed to exploit attribute information for effective V-ReID. In previous works that use attributes, there is a lack of interaction between the global V-ReID head and the attribute regression heads, which makes that the feature information is not effectively exploited for V-ReID.

To address this issue, we propose ANet (as shown in Figures 1 and 2). It consists of two parts: VAD based Network (VAN) and Joint Module (JM). VAN is based on a Backbone with two heads, where one of them is to learn global V-ReID features and the other to regress attributes. VAN outputs an initial feature representation of V-ReID and multiple Attribute features from the input image. Then, the JM distills V-ReID-helpful attribute information and compensates it into the global features. JM promotes the interaction between the two heads of VAN. Furthermore, we propose an Amelioration Constraint (AC), which drives the attribute compensated feature to be more discriminative than the original V-ReID feature before the compensation.

### 3.1. VAD based Network

VAD based Network (VAN), shown in Figure 1, aims to learn V-ReID features and regress attributes. This design is similar to previous literature work, where the attribute branches are expected to drive the learning of robust features since the attributes are in general invariant to illumination and viewpoint changes.

**Backbone.** A backbone network is used to extract feature map $F(I) \in \mathbb{R}^{h \times w \times c}$ from an input image $I$, where $h$, $w$ and $c$ are height, width and channels of $F(I)$, respectively. We follow the previous works and use ResNet [16] as the backbone.

**V-ReID Head/Branch.** On top of the backbone feature $F(I)$, we append a spatial global average pooling (GAP) layer followed by a fully-connected (FC) layer to generate the V-ReID feature $f(I)$ as

$$f(I) = W_f \, \text{GAP}\,(F(I)) + \mathbf{b_f}, \quad (1)$$

where $W_f$ and $\mathbf{b_f}$ denote the weights and bias of the FC layer used to reduce the dimension of the pooled feature, $W_f \in \mathbb{R}^{s_f \times c}$ and $\mathbf{b_f} \in \mathbb{R}^{s_f}$, where $s_f$ is the predefined dimension of the output. $f(I)$ is followed by Triplet Loss $L_{tri}^f$ and Cross Entropy Loss $L_{ID}^f$.

**Attribute Heads/Branches.** On top of the backbone feature $F(I)$, we add $n$ attribute branches for attribute classification, where $n$ is the number of available attributes in the training dataset, one branch for each attribute. For the $i$-th attribute branch, we use a spatial and channel attention module to obtain attribute-related feature $A_i(I) \in \mathbb{R}^{h \times w \times c}$ as

$$A_i(I) = F(I) \cdot Att_i(F(I)), \quad (2)$$

where $Att_i(I) \in \mathbb{R}^c$ denotes the response of the attention module.

To make classification for the $i$-th attribute, we apply GAP and a FC layer to get a feature vector $a_i$ as

$$a_i(I) = W_{a_i} \text{GAP}(A_i(I)) + \mathbf{b_{a_i}}, \quad (3)$$

where $W_{a_i}$ and $\mathbf{b_{a_i}}$ denote the weights and bias of the FC layer, $W_{a_i} \in \mathbb{R}^{s_a \times c}$ and $\mathbf{b_{a_i}} \in \mathbb{R}^{s_a}$, where $s_a$ is the predefined size of the output. $a_i(I)$ is followed by a classifier with a cross entropy loss $L_{att}^i$ to recognize which class it belongs to for the $i$-th attribute.

In summary, VAN is trained by minimizing the loss $L_{VAN}$ as

$$L_{VAN} = L_{tri}^f + L_{ID}^f + \lambda_A \sum_{i=1}^{n} L_{att}^i, \quad (4)$$

where $\lambda_A$ is a hyper-parameter for balancing the importance of V-ReID loss and attribute-related losses.

### 3.2. Joint Module

The Joint Module (JM) is illustrated in Figure 2. JM aims to distill V-ReID helpful information from the attribute features and compensate it to the V-ReID feature for the final feature matching. First, we merge the attribute feature maps from multiple branches to have a unified attribute feature map $G(I)$. Then, we distill discriminative V-ReID helpful information from $G(I)$ and compensate it onto $F(I)$ to create a Joint Feature $J(I)$. To encourage a higher discriminative capability of the Joint Feature, we propose an Amelioration Constraint (AC).

**Attribute Feature** $G(I)$. To facilitate the distillation of helpful attribute features, we combine all the attribute feature maps $A_i(I)$, where $i = 1, \cdots, n$, to have a unified attribute feature map $G(I)$. We achieve this by summarizing the attribute feature maps followed by a convolution layer and a residual connection as

$$G(I) = \sum_{i=1}^{n} A_i(I) + \theta_A(\sum_{i=1}^{n} A_i(I)), \quad (5)$$

where $\theta_A$ is implemented by a $1 \times 1$ convolutional layer followed by batch normalization (BN) and ReLU activation, that is, $\theta_A(\mathbf{x}) = \text{ReLU}(W_A \mathbf{x})$, $W_A \in \mathbb{R}^{c \times c}$. We omit BN to simplify the notation.

For the combined attribute feature map $G(I)$, we add supervision from attributes to preserve the attribute information. Given $n$ attributes, $m_i$ is the number of classes for the $i$-th attribute. There are in total $\prod_{i=1}^{n} m_i$ attribute patterns. We apply a GAP layer on $G(I)$ to get the feature vector $g(I)$. Then, the Triplet Loss $L_{tri}^g$ is used as supervision to pull the features for the same attribute pattern and push the features for the different attribute patterns. We name this supervision as Attribute-based Triplet Loss.

**Joint Feature** $J(I)$. To distill V-ReID-helpful attribute information from $G(I)$ to enhance $F(I)$, we use two convolution layers to have distilled feature $G_{reid}(I)$

$$G_{reid}(I) = \theta_{g1}(\theta_{g2}(G(I))), \quad (6)$$

where $\theta_{g1}$ and $\theta_{g2}$ are implemented similarly to $\theta_A$ but we use a $3 \times 3$ convolutional layer instead of $1 \times 1$, $\theta_{g1}(\mathbf{x}) = \text{ReLU}(W_{g1}\mathbf{x})$, $\theta_{g2}(\mathbf{x}) = \text{ReLU}(W_{g2}\mathbf{x})$, $W_{g1} \in \mathbb{R}^{c \times c}$ and $W_{g2} \in \mathbb{R}^{c \times c}$.

By adding $G_{reid}(I)$ onto the V-ReID feature $F(I)$, we have the Joint Feature $J(I)$ as

$$J(I) = F(I) + G_{reid}(I). \quad (7)$$

$J(I)$ combines V-ReID information from $F(I)$ and the relevant V-ReID-helpful information from the attributes $G(I)$. Similar to the supervision on $F(I)$, we add Triplet Loss $L_{tri}^j$ and Cross Entropy Loss $L_{ID}^j$ on the spatially average pooled feature $j(I)$, where $j(I)$ is obtained as

$$j(I) = W_j \cdot \text{GAP}(J(I)) + \mathbf{b_j}, \quad (8)$$

where $W_j$ and $\mathbf{b_j}$ represent the weights and bias of a FC layer, $W_j \in \mathbb{R}^{s_j \times c}$ and $\mathbf{b_j} \in \mathbb{R}^{s_j}$, $s_j$ is the predefined dimension of the output. JM is trained by minimizing $L_{JM}$

$$L_{JM} = L_{tri}^j + L_{ID}^j + \lambda_G L_{tri}^g, \quad (9)$$

where $\lambda_G$ is a hyperparameter balancing the importance of the compensated V-ReID loss and the attribute related loss.

Finally, we can train the entire network ANet end-to-end by minimizing $L$

$$L = L_{JM} + \lambda L_{VAN}, \quad (10)$$

where $\lambda$ is a hyperparameter to balance the importance of $L_{JM}$ and $L_{VAN}$.

**Amelioration Constraint.** To further boost the capabilities of the network, we define the Amelioration Constraint (AC). AC aims to explicitly push $j(I)$ to be more discriminative than $f(I)$. We separately apply AC for cross entropy loss and triplet loss.

*AC for Cross Entropy Loss:* For image $I$, we define it as

$$AC_{ID}(I) = \text{softplus}(L_{ID}^j(I) - L_{ID}^f(I)), \quad (11)$$

where $\text{softplus}(\cdot) = \ln(1 + \exp(\cdot))$ is a monotonically increasing function that helps to reduce the optimization difficulty by avoiding negative values [26]. $L_{ID}^f(I)$ and $L_{ID}^j(I)$ represent the identity cross entropy loss with respect to feature $f(I)$ and $j(I)$, respectively. Minimizing $AC_{ID}(I)$ encourages the network to have a lower classification error for $j(I)$ than that for $f(I)$.

*AC for Triplet Loss:* We seek $j(I)$ to represent an enhanced feature of $f(I)$, where $j(I)$ has a higher discriminative capability than $f(I)$. Thus, we encourage the feature distance $D(\cdot,\cdot)$ between an anchor sample/image $I$ and a positive sample $I^+$ to be smaller w.r.t. feature $j(\cdot)$ than feature $f(\cdot)$. Similarly, we encourage the feature distance $D(\cdot,\cdot)$ between an anchor sample/image $I$ and a negative sample $I^-$ to be larger w.r.t. feature $j(\cdot)$ than feature $f(\cdot)$. Then, AC for triplet loss $AC_{tri}$ is defined as

$$AC_{tri}(I) = \text{softplus}(D(j(I), j(I^+)) - D(f(I), f(I^+))) + \\ \text{softplus}(D(f(I), f(I^-)) - D(j(I), j(I^-))). \quad (12)$$

We notice that training with $AC_{ID}, AC_{tri}$ in an end-to-end leads to unstable learning. Thus, we follow two steps in training. In the first step, we minimize $L$. In the second step, we freeze the backbone (that is, all operations before $f$) and minimize $L'$. Compared with $L$ in Equation (10), the AC losses are enabled and the losses on feature $f$ are disabled in $L'$ as

$$L' = L + AC_{tri} + AC_{ID} - \lambda(L_{tri}^f + L_{ID}^f). \quad (13)$$

## 4. Experiments

In this section, we present the datasets used in our experiments, the implementation details, an ablation study and a comparison against the state of the art to validate our proposed method.

### 4.1. Datasets

We evaluate our vehicle re-identification method on three challenging benchmark datasets.

- **VeRi776** [27]: It contains over 50,000 images of 776 vehicles with 20 camera views. It includes attribute labels for color and type. It considers 576 vehicles for training and 200 vehicles for test.

- **VeRi-Wild** [28]: This is the largest vehicle re-identification dataset. It considers 174 camera views, 416,314 images and 40,671 IDs. It includes attribute labels for vehicle model, color and type. The testing set is divided into three sets with 3,000 (small), 5,000 (medium) and 10,000 (large) IDs. This is the most challenging dataset because the images were captured for a period of one month and include severe changes in background, illumination, viewpoint and occlusions.

- **Vehicle-ID** [15]: It includes 221,763 images of 26,267 vehicles, captured from either front or back views. The training set contains 110,178 images of 13,134 vehicles and the test set contains 111,585 images of 13,133 vehicles. The testing data is further divided into three sets with 200 (small), 1,600 (medium) and 2,400 (large) vehicles. Some images in this dataset have attribute labels for vehicle color and type but not for all the images.

For the first two datasets, the validation protocols is based on mean Average Precision (mAP) and Cumulative Matching Curve (CMC) @1 (at rank-1/R1) and @5 (at rank-5/R5) as they have fixed gallery and query sets. For Vehicle-ID, we follow the protocol proposed by the authors of the dataset, which randomly chooses one image of each vehicle ID as gallery and the rest as query. The final R1 and R5 results are reported after repeating this process 10 times.

### 4.2. Implementation Details

We follow other works in the literature to implement the backbone for a fair comparison. We use a modified version of ResNet-50 [16] with Instance-Batch Normalization (IBN) [29] and remove the last pooling layer to obtain the feature map $F(I)$ for an image $I$. Each attention module $Att_i(I)$ is based on SE [30] with the reduction ratio of 16. For the FC layers, we set $s_a = 128$ and $s_f = s_j = 512$.

We use cross entropy loss with label smoothing (LS) regularization [31] and triplet loss with hard positive-negative mining [32], following the Bag-of-Tricks [33].

In one of the datasets, not all input images have attribute labels. For these samples, we simply do not backpropagate the losses from $L_{att}^i$ and $L_{tri}^g$. We found this works well since we use batch size of 512 (4 images per ID) and the missing labels are alleviated by the other IDs in the batch. Note that these missing labels do not affect our $AC_{ID}$ and $AC_{tri}$, so ANet can still learn from those cases.

The input images are resized to 256×256 pixels and augmented by random horizontal flipping, random zooming and random input erasing [34, 35, 36, 37]. All models are trained on 8 V100 GPUs with NVLink for 210 epochs with Amsgrad. An initial learning rate is set to 0.0006 and the learning rate is decayed by 0.1 at epochs 60, 120 and 150. The first learning step minimizes $L$ for the first 150 epochs, then the second step optimizes $L'$ for 60 epochs. $n = 2$ for all datasets, where we consider vehicle color (for instance, red, yellow, green) and type (such as sedan and truck). During testing, the feature vectors are L2-normalized for matching.

Table 1: Ablation study on the effectiveness of our designs. We indicate the feature vector used for testing using the symbol in parenthesis.

| | VeRi776 | | Vehicle-ID | | | | | | VeRi-Wild | | | | | |
| | | | Small | | Medium | | Large | | Small | | Medium | | Large | |
| Method | mAP | R1 | R1 | R5 | R1 | R5 | R1 | R5 | mAP | R1 | mAP | R1 | mAP | R1 |
| --- | --- | --- | --- | --- | --- | --- | --- | --- | --- | --- | --- | --- | --- | --- |
| Baseline | 78.1 | 96.1 | 81.3 | 94.4 | 77.7 | 90.6 | 75.8 | 88.5 | 78.1 | 94.6 | 72.2 | 92.5 | 64.0 | 88.7 |
| VAN ($f$) | 78.1 | 96.6 | 84.1 | 96.5 | 80.4 | 93.6 | 78.4 | 91.8 | 83.1 | 94.5 | 78.3 | 93.5 | 70.6 | 90.0 |
| VAN ($fa$) | 77.3 | 96.5 | 81.5 | 95.0 | 78.5 | 92.0 | 76.3 | 89.6 | 81.9 | 94.1 | 76.9 | 93.1 | 69.2 | 89.4 |
| ANet ($j$) w/o AC | 79.8 | 96.9 | 85.0 | 96.7 | 80.9 | 94.1 | 79.0 | 91.8 | 84.6 | 96.1 | 79.9 | 94.4 | 72.9 | 91.5 |
| ANet ($j$) | 80.1 | 96.9 | 86.0 | 97.4 | 81.9 | 95.1 | 79.6 | 92.7 | 85.8 | 95.9 | 81.0 | 94.5 | 73.9 | 91.6 |

*4.3. Ablation Study*

Our ablation study contains four subsections. In the first two subsections, we analyze the effectiveness of our ANet and its components (i.e., Joint Module and AC). In the third subsection, we aim to analyze the design of previous methods using attributes represented by VAN. Specifically, we analyze the effects of using attributes to define attention masks, instead of the FC layers, which are common in existing works. The final subsection studies the influence of hyperparameters $\lambda_A$, $\lambda_G$ and $\lambda$, as well as IBN and LS.

*4.3.1. Effectiveness of using Attributes on V-ReID*

We first evaluate the effects of using attributes in V-ReID and show the comparisons in Table 1. *Baseline* denotes the scheme which generates feature $f$ using only the backbone, without using attribute-related designs. *VAN* denotes the vanilla scheme that explores attributes as shown in Figure 1, using the same backbone as *Baseline*.

For our VAN, we can use the V-ReID feature $f(i)$ (i.e., *VAN ($f$)*), or use the concatenation of $f(I)$ and attribute features $a_i(I), i = 1, \cdots, n$ (i.e., *VAN ($fa$)*) in inference. We can see that: 1) *VAN ($f$)*, where the attributes regularize the feature learning, outperforms *Baseline* significantly on Vehicle-ID and VeRi-Wild. Specially, using attributes improves the rank-1 by 0.5% for VeRi776, 2.8% at rank-1 and 3.3% at rank-5 for Vehicle-ID, 6.6% in mAP and 1.3% at rank-1 for VeRi-Wild; 2) using *VAN ($fa$)* has lower performance than *VAN ($f$)*. This is because not all the attribute information $a_i(I)$ is equally important for V-ReID. Allocating the relative contributions of each attribute is needed to have satisfactory results. Hence how to distill task-oriented attribute information to efficiently benefit V-ReID is important, which is what our ANet aims to address.

*4.3.2. ANet: A Superior Way to Distill Attributes Information*

We propose ANet to distill attribute information for more effective V-ReID. Here we study the effectiveness of our Joint Module design, and the AC losses. Table 1 shows the comparisons. We can see that: (i) Our final scheme *ANet ($j$)* significantly outperforms the basic network *VAN ($f$)*, by 2.0% in mAP on VeRi776, 1.9%/1.5%/1.5% in Rank-1 on Small/Medium/Large scales of Vehicle-ID, 2.7%/2.7%/3.3% in mAP on Small/Medium/Large scales of VeRi-Wild; (ii) our proposed AC losses generates higher discrimination after the compensation of distilled attribute feature than that before and is very helpful to promote the distill of discriminative information from attribute feature for V-ReID purpose.

These results show that the interaction between the V-ReID and attribute features of VAN improves the network performance, thanks to the distill of V-ReID oriented attribute features.

To better understand the effects of ANet, we visualize the attention maps of $G(I)$ and $G_{reid}(I)$ and show some in Figure 3. $G(I)$ encodes generic features of the attributes, where the activations are flatter and do not have a special focus on the vehicle parts. In contrast, $G_{reid}(I)$ represents a portion of the information of $G(I)$ that is helpful for V-ReID. We can observe that the activation maps focus more on the vehicle.

To further analyze the effectiveness of our proposed ANet, we compare our interaction design with that using attributes as attention, which we refer to as ANet (att). In ANet (att), attention is learned based on attributes using CBAM [38] and is used to generate attribute-guided features similar to AGNet [8]. In this

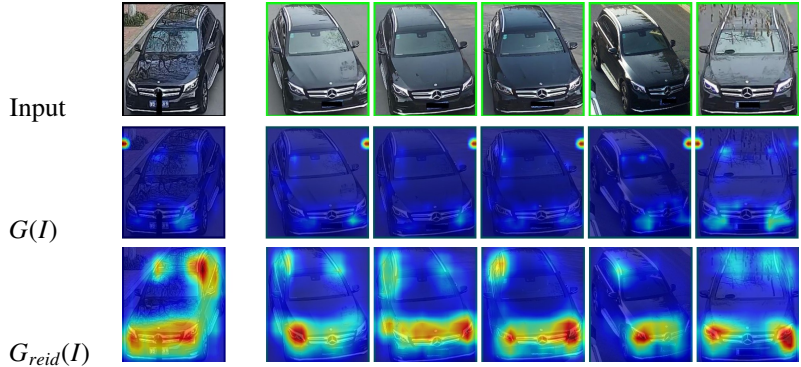

Figure 3: Comparison of activation maps. The first row represents the input images, second and third row their corresponding activation maps for $G(I)$ (attribute features) and $G_{reid}(I)$ (attribute features oriented to V-ReID), respectively. The first column is the query image, the second to sixth columns represent the vehicle retrieved at rank-1, rank-2, rank-3, rank-4 and rank-5.

case, $J(I)$ is defined through Equation (14) as

$$J(I) = F(I) \cdot CBAM(G(I)) + F(I). \qquad (14)$$

Table 2: Comparison of our interaction design with that using attributes as attention. Note that the results for Vehicle-ID and VeRi-Wild are reported using their small scale test set.

| | VeRi776 | | Vehicle-ID | | VeRi-Wild | |
|---|---|---|---|---|---|---|
| Method | mAP | R1 | R1 | R5 | mAP | R1 |
| Baseline | 78.1 | 96.1 | 81.3 | 94.4 | 78,1 | 94.6 |
| ANet (att) | 78.2 | 96.1 | 83.9 | 96.2 | 84.9 | 95.5 |
| ANet | 80.1 | 96.9 | 86.0 | 97.4 | 85.8 | 95.9 |

We compare the performance of ANet (att) with our Anet in Table 2. We can see that using attention directly to increase the interaction between attribute and ReID heads is not as effective as ours. Distilling the ReID-relevant information from the attribute head defined by $G_{reid}$ provides a superior performance. Furthermore, ANet (att) has a performance similar to the simple baseline VAN.

### 4.3.3. VAN: Attention vs Fully Connected

We use VAN as our *attribute-based* baseline, which is similar to previous works that explore vehicle attributes. However, previous works commonly used simple FC layers, instead of attention blocks for the attribute branches. Using attention facilitates the distillation of attribute features. As shown in Table 3, attention outperforms the use of FC layers by 1.2% in rank-1 on Vehicle-ID, as well as 1.4% and 1% in mAP on VeRi776 and VeRi-Wild datasets, respectively.

Table 3: Comparison of choice for implementation of attribute branches for the attribute-based baseline VAN. *fc* represents an implementation using fully connected layers and *att* represents an implementation based on SE attention blocks. Results for Vehicle-ID and VeRi-Wild are reported using their small scale test set.

| | VeRi776 | | Vehicle-ID | | VeRi-Wild | |
|---|---|---|---|---|---|---|
| Method | mAP | R1 | R1 | R5 | mAP | R1 |
| fc | 76.7 | 95.8 | 83.3 | 96.0 | 82.1 | 94.3 |
| att | 78.1 | 96.6 | 84.1 | 96.5 | 83.1 | 94.5 |

### 4.3.4. Hyperparameter Analysis

Both hyperparameters $\lambda_A$ (in Equation (4) and $\lambda_G$ (in Equation (9)) balance the importance of the attribute information in the total loss. We study their influence and show the results in Table 4. We can see that assigning the same weight for both attribute and V-ReID signals (e.g., $\lambda_A = \lambda_G = 1$) provides the best results.

Interestingly, assigning a low weight to the attribute signals (e.g., $\lambda_A = 0.01$ and $\lambda_G = 0.01$) decreases their impact and results in inferior performance. Furthermore, giving high weights to attribute signal (e.g., $\lambda_A = 100$ and $\lambda_G = 100$) is better than assigning them with rather low weights. This shows the importance of the attribute information in our pipeline for V-ReID. Finally, $\lambda$ (in Equation (10) balances the importance of VAN and JM.

We observe that a high weight to VAN (e.g., $\lambda = 100$) significantly decreases the performance, where the contribution of our JM is small. The best weight to combine VAN and JM is $\lambda = 1$. Based on this analysis, we set $\lambda_A = \lambda_G = \lambda = 1$ and use these values in the remaining experiments.

In both our baseline scheme and final scheme, we fol-low the common practice and use Instance Batch Normalization (IBN) and Label Smoothing (LS). Here, we study the influence of IBN and LS on the performance of our ANet and show the results in Tables 5 and 6, respectively.

Table 4: Ablation study on the influence of $\lambda_A$, $\lambda_G$ and $\lambda$. We evaluate using VeRi776 by keeping the non-tested hyperparameters fixed. For example, in order to analyze $\lambda_A$, we set $\lambda_G = 1, \lambda = 1$.

| Results | $\lambda_A$ 0.01 | 0.1 | 1.0 | 10.0 | 100.0 |
|---|---|---|---|---|---|
| mAP | 59.1 | 79.4 | 80.1 | 77.5 | 65.9 |
| R1 | 88.6 | 96.5 | 97.1 | 96.2 | 91.4 |
| R5 | 94.7 | 98.6 | 98.6 | 98.4 | 96.1 |
| Results | $\lambda_G$ 0.01 | 0.1 | 1.0 | 10.0 | 100.0 |
| mAP | 60.3 | 61.7 | 80.1 | 76.7 | 68.1 |
| R1 | 90.0 | 90.9 | 97.1 | 95.8 | 91.6 |
| R5 | 95.2 | 95.5 | 98.6 | 98.6 | 96.7 |
| Results | $\lambda$ 0.01 | 0.1 | 1.0 | 10.0 | 100.0 |
| mAP | 78.3 | 79.7 | 80.1 | 64.4 | 56.4 |
| R1 | 96.8 | 96.4 | 97.1 | 92.1 | 87.9 |
| R5 | 98.3 | 98.2 | 98.6 | 96.3 | 94.5 |

For IBN, we can observe a considerable decrease of 1.7% in mAP on VeRi776 when not using IBN, whereas a decrease of 0.7% in R1. For Vehicle-ID, the difference is also significant, a decrease of 2.4% for R1 and 0.5% for R5. For VeRi-Wild, not using IBN has a smaller effect than on other datasets, that is, 0.2% in both R1 and mAP.

Table 5: Influence of Instance Batch Normalization (IBN). Results for Vehicle-ID and VeRi-Wild are reported using their small scale test set.

| | VeRi776 | | Vehicle-ID | | VeRi-Wild | |
|---|---|---|---|---|---|---|
| Method | mAP | R1 | R1 | R5 | mAP | R1 |
| Baseline w/o IBN | 69.5 | 91.0 | 71.5 | 83.1 | 69.2 | 89.2 |
| Baseline | 78.1 | 96.1 | 81.3 | 94.4 | 78,1 | 94.6 |
| ANet w/o IBN | 78.4 | 96.2 | 84.4 | 96.9 | 85.6 | 95.7 |
| ANet | 80.1 | 96.9 | 86.0 | 97.4 | 85.8 | 95.9 |

Without using LS, there is a decrease of 1.1% in mAP on VeRi776, 1.2% in R1 on Vehicle-ID, 0.1% in mAP on VeRi-Wild, respectively. In general, we can observe that IBN has a more significant importance than LS in the final performance.

To analyze each of the weights $\lambda_A$, $\lambda_G$ and $\lambda$ for our loss functions, we conduct an independent analysis per parameter (for instance, in order to analyze $lambda_A$, we evaluate 5 different values for it and fix $\lambda_G = \lambda = 1$). Results are shown in Table 4.

Table 6: Influence of Label Smoothing (LS). Results for Vehicle-ID and VeRi-Wild are reported using their small scale test set.

| | VeRi776 | | Vehicle-ID | | VeRi-Wild | |
|---|---|---|---|---|---|---|
| Method | mAP | R1 | R1 | R5 | mAP | R1 |
| Baseline w/o LS | 73.2 | 93.2 | 73.8 | 88.4 | 74.3 | 90.1 |
| Baseline | 78.1 | 96.1 | 81.3 | 94.4 | 78,1 | 94.6 |
| ANet w/o LS | 79.0 | 96.8 | 85.8 | 97.2 | 85.7 | 95.5 |
| ANet | 80.1 | 96.9 | 86.0 | 97.4 | 85.8 | 95.9 |

### 4.4. Comparison with State-of-the-Art Methods

We compare our method with approaches that also use attributes information [6, 40, 7, 8, 9, 10]. We also compare our method with the most recent works that leverage clue/approach such as vehicle parsing maps [5], vehicle parts [12, 20], GANs [18], Teacher-Student (TS) distillation [41, 42], camera viewpoints [11, 42], and Graph Networks (GN) [19, 13]. HPGN creates a pyramid of spacial graph network to explore the spatial significance of the backbone tensor. PCRNet studies the correlation between parsed vehicle parts through a graph network. VAnet [11] learns two metrics for similar viewpoints and different viewpoints in two feature spaces, respectively.

We also compare against FastReid [43], a strong baseline network for re-identification that performs an extensive search of hyperparameters, augmentation methods, and use some architecture design tricks to achieve excellent performance. We also implemented our design on top of it by taking it as our backbone, which we named *ANet + FastReid*. Note that the reported results of FastReid were obtained by our running of their released code. Tables 7, 8 and 9 show the comparisons on VeRi776, Vehicle-ID, and VeRi-Wild, respectively.

**VeRi776**. Compared with attribute-based methods (first group in Table 7), our scheme *ANet + FastReid* outperforms the best results in this group by **5.1%** in mAP; and 1.5% for rank-1 and rank-5. By comparing with methods that do not use attributes, we can see that it performs the second best in mAP, and achieves the best for rank-1 and rank-5. VKD [42] is better than ours in mAP and is inferior to ours at rank-1 and rank-5, where VKD uses camera labels in training to be viewpoint-invariant and trains a model based on the Teacher-Student framework.

Table 7: Comparison of our proposed method against the state of the art on VeRi776. The first and second best results are marked by **bold** and underline, respectively.

| Method | Clue/Approach | mAP | R1 | R5 |
|---|---|---|---|---|
| PAMAL [39] | attributes | 45.0 | 72.0 | 88.8 |
| MADVR [40] | attributes | 61.1 | 89.2 | 94.7 |
| DF-CVTC [6] | attributes | 61.0 | 91.3 | 95.7 |
| PAMTRI [7] | attributes | 71.8 | 92.8 | 96.9 |
| AGNet [8] | attributes | 71.5 | 95.6 | 96.5 |
| SAN [9] | attributes | 72.5 | 93.3 | 97.1 |
| StRDAN [10] | attributes | 76.1 | – | – |
| VAnet [11] | viewpoint | 66.3 | 89.7 | 95.9 |
| PRND [20] | veh. parts | 74.3 | 94.3 | 98.6 |
| UMTS [41] | TS | 75.9 | 95.8 | – |
| PCRNet [13] | GN + parsing | 78.6 | 95.4 | 98.4 |
| SAVER [18] | GAN | 79.6 | 96.4 | 98.6 |
| PVEN [5] | parsing | 79.5 | 95.6 | 98.4 |
| HPGN [19] | GN | 80.1 | 96.7 | – |
| VKD [42] | viewpoint + TS | **82.2** | 95.2 | 98.0 |
| Baseline | attributes | 78.1 | 96.1 | 98.3 |
| ANet (Ours) | attributes | 80.1 | **97.1** | **98.6** |
| FastReid [43] | backbone | 81.0 | 97.1 | 98.3 |
| ANet + FastReid (Ours) | attributes | <u>81.2</u> | 96.8 | 98.4 |

Table 8: Comparison of our proposed method against the state of the art on Vehicle-ID. The first and second best results are marked by **bold** and underline, respectively.

| | | Small | | Medium | | Large | |
|---|---|---|---|---|---|---|---|
| Method | Clue/Approach | R1 | R5 | R1 | R5 | R1 | R5 |
| PAMAL [39] | attributes | 67.7 | 87.9 | 61.5 | 82.7 | 54.5 | 77.2 |
| AGNet [8] | attributes | 71.1 | 83.7 | 69.2 | 81.4 | 65.7 | 78.2 |
| DF-CVTC [6] | attributes | 75.2 | 88.1 | 72.1 | 84.3 | 70.4 | 82.1 |
| SAN [9] | attributes | 79.7 | 94.3 | 78.4 | 91.3 | 75.6 | 88.3 |
| PRND [20] | veh. parts | 78.4 | 92.3 | 75.0 | 88.3 | 74.2 | 86.4 |
| SAVER [18] | GAN | 79.9 | 95.2 | 77.6 | 91.1 | 75.3 | 88.3 |
| UMTS [41] | TS | 80.9 | – | 78.8 | – | 76.1 | – |
| PVEN [5] | parsing | 84.7 | 97.0 | 80.6 | 94.5 | 77.8 | 92.0 |
| PCRNet [13] | GN + parsing | 86.6 | **98.1** | 82.2 | **96.3** | 80.4 | 94.2 |
| VAnet [11] | viewpoint | 88.1 | 97.2 | **83.1** | 95.1 | 80.3 | 92.9 |
| HPGN [19] | GN | **89.6** | – | 79.9 | – | 77.3 | – |
| Baseline | attributes | 81.3 | 94.4 | 77.7 | 90.6 | 75.8 | 88.5 |
| ANet (Ours) | attributes | 86.0 | 97.4 | 81.9 | 95.1 | 79.6 | 92.7 |
| FastReid [43] | backbone | 85.5 | 97.4 | 81.8 | 95.3 | 79.9 | 93.8 |
| ANet + FastReid (Ours) | attributes | 87.9 | <u>97.8</u> | <u>82.8</u> | <u>96.2</u> | **80.5** | **94.6** |

**Vehicle-ID**. Our method outperforms attribute-based methods (first group in Table 8) consistently. For rank-1, our scheme *ANet + FastReid* outperforms the best attribute-based method by **8.2%**, **4.4%** and **4.9%** for small, medium and large scales, respectively. When compared with methods using other clues, ours achieves in the best results on the large set and competitive performance on the other sets.

**VeRi-Wild**. Previous attribute based methods have not yet reported results for this latest dataset. From Table 9, we can see that our schemes *ANet* and *ANet + FastReid* achieve the best performance in mAP.

Table 9: Comparison of our proposed method against the state of the art on VeRi-Wild. The first and second best results are marked by **bold** and underline, respectively.

|  |  | Small | | | Medium | | | Large | | |
|---|---|---|---|---|---|---|---|---|---|---|
| **Method** | **Clue/Approach** | **mAP** | **R1** | **R5** | **mAP** | **R1** | **R5** | **mAP** | **R1** | **R5** |
| UMTS [41] | TS | 82.8 | 84.5 | – | 66.1 | 79.3 | – | 54.2 | 72.8 | – |
| HPGN [19] | GN | 80.4 | 91.3 | – | 75.1 | 88.2 | – | 65.0 | 82.6 | – |
| PCRNet [13] | GN + parsing | 81.2 | 92.5 | – | 75.3 | 89.6 | – | 67.1 | 85.0 | – |
| SAVER [18] | GAN | 80.9 | 94.5 | 98.1 | 75.3 | 92.7 | 97.4 | 67.7 | 89.5 | 95.8 |
| PVEN [5] | parsing | 82.5 | **96.7** | 99.2 | 77.0 | **95.4** | **98.8** | 69.7 | **93.4** | **97.8** |
| Baseline | attributes | 78.1 | 94.6 | 98.5 | 72.2 | 92.5 | 97.3 | 64.0 | 88.7 | 95.6 |
| ANet (Ours) | attributes | 85.8 | 95.9 | 99.0 | 81.0 | 94.5 | 98.1 | 73.9 | 91.6 | 96.7 |
| FastReid [43] | backbone | 84.8 | 95.7 | 98.9 | 80.0 | 94.5 | 98.1 | 73.2 | 91.5 | 96.7 |
| ANet + FastReid (Ours) | attributes | **86.9** | <u>96.5</u> | **99.2** | **82.5** | <u>95.2</u> | <u>98.3</u> | **75.9** | <u>92.5</u> | <u>97.2</u> |

PVEN [5] is a method based on semantic parsing to describe each vehicle view and region. It has better results on rank-1/rank-5 but it is not as competitive as in the two previous datasets.

We observed that none of the existing methods consistently achieve the best results on all the datasets. This may be because different datasets have different main challenges. Our proposed ANet shows a more consistent state-of-the-art performance on all the datasets, thanks to the generic capabilities of attributes on V-ReID.

## 5. Conclusions

In this work, we proposed ANet, a novel framework to leverage attribute information for vehicle re-identification. ANet addresses the problem of lack of interaction between the V-ReID features and attribute features of previous methods. Particularly, we encourage the network to distill task-oriented information from the attribute branches and compensate it into the global V-ReID feature to enhance the discrimination capability of the feature. Evaluation on three datasets shows the effectiveness of our methods.

## Acknowledgments

This work was done while the first author is affiliated with Microsoft Corp. We are thankful to Microsoft Research, São Paulo Research Foundation (FAPESP grant #2017/12646-3), National Council for Scientific and Technological Development (CNPq grant #309330/2018-1) and Coordination for the Improvement of Higher Education Personnel (CAPES) for their financial support.